\documentclass[preprint,12pt]{elsarticle}

\usepackage{color}
\usepackage{amsmath,amssymb}
\usepackage{framed}
\usepackage{graphicx}
\usepackage{amssymb}
\usepackage{lineno}
\usepackage{hyperref}
\usepackage{url}
\usepackage{multirow}
\usepackage{verbatim}
\usepackage{bm}
\usepackage[linesnumbered,ruled]{algorithm2e}

\journal{International Journal of Medical Informatics}

\begin{document}

\begin{frontmatter}

\newcommand{\tr}[1]{\textcolor{red}{#1}}


\title{A Methodology for Customizing Clinical Tests for Esophageal Cancer based on Patient Preferences}

\author{Asis Roy\fnref{myfootnote1}\corref{mycorrespondingauthor}}
\ead{asisry@gmail.com}
\author{Sourangshu Bhattacharya\fnref{myfootnote2}}
\author{Kalyan Guin\fnref{myfootnote3}}
\address{Indian Institute of Technology(IIT), Kharagpur, India}
\fntext[myfootnote1]{Research Scholar.}
\fntext[myfootnote2]{Assistant Professor,Computer Science Dept.IIT, Kharagpur.}
\fntext[myfootnote3]{Dean and Professor, VGSOM, IIT, Kharagpur.}

\cortext[mycorrespondingauthor]{Corresponding author}

\begin{abstract}

\textbf{Background: }
Tests for Esophageal cancer, e.g. the Barium swallow test, can be expensive, uncomfortable and can have side effects. For many patients, we can predict non-existence of disease with 100\% certainty, just using demographics, lifestyle, and medical history information. Our objective is to devise a general methodology for customizing tests using user preferences so that expensive or uncomfortable tests can be avoided.


\textbf{Method: }
We propose to use classifiers trained from electronic health records (EHR) for selection of tests. The key idea is to design classifiers with 100\% false normal rates, possibly at the cost higher false abnormals. We compare Naive Bayes classification (NB), Random Forests (RF), Support Vector Machines (SVM) and Logistic Regression (LR), and find kernel Logistic regression to be most suitable for the task. We propose an algorithm for finding the best probability threshold for kernel LR, based on test set accuracy. Using the proposed algorithm, we describe schemes for selecting tests, which appear as features in the automatic classification algorithm, using preferences on costs and discomfort of the users.


\textbf{Result: }
We test our methodology with EHRs collected for more than $3000$ patients, as a part of project carried out by a reputed hospital in Mumbai, India.
Naive versions of NB, RF, SVM and LR provide good accuracy and sensitivity with medical practitioner (MP) observations included in the features, but show very poor sensitivity ($\sim 40\%$) without them. We show that kernel SVM and kernel LR with a polynomial kernel of degree $3$, yields an accuracy of $99.8\%$ and sensitivity $100\%$, without the MP features, i.e. using only clinical tests. We demonstrate our test selection algorithm using two case studies, one using cost of clinical tests, and other using "discomfort" values for clinical tests. We compute the test sets corresponding to the lowest false abnormals for each criterion described above, using exhaustive enumeration of $15$ clinical tests. The sets turn out to different, substantiating our claim that one can customize test sets based on user preferences.


\end{abstract}


\end{frontmatter}


\section{Introduction}
\label{S:1}


Personalized medicine (PM) is a field of healthcare which finds the best available diagnosis or tailored diagnosis to satisfy the need of an individual patient. According to Barrack Obama, it gives us "one of the greatest opportunities for new medical breakthroughs that we have ever seen" \cite{obama}.
Healthcare providers can build a revolutionary new system for medical care by combining the data from diagnostic tests and medical history of patients to deliver enhanced value to patients. One such possibility is \textit{personalized tests}, where the set of clinical tests recommended to a patient for diagnosis of a disease, are customized according to the patients needs. In this paper, we study the problem of designing such a system and propose a methodology for it.

\begin{figure}[h]
\centering\includegraphics[width=.8\linewidth]{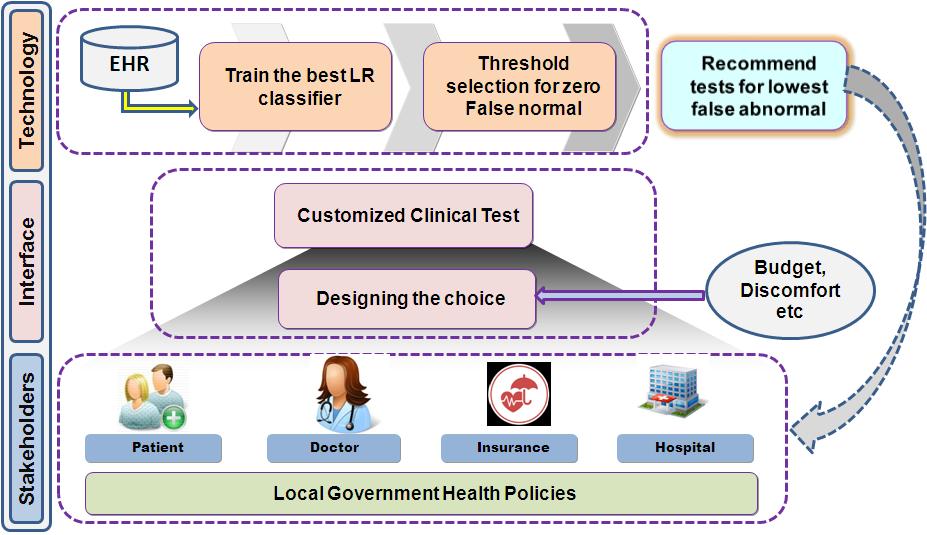}
\caption{Overview of Personalized Diagnosis System}
\label{fig:overview}
\end{figure}

A central idea behind most studies in personalized medicine is to design drugs specific to a patients genetic predisposition. This is expected to work especially well, when the drug is for a disease due to genetic abnormality, e.g. various forms of cancer and other inherited diseases. However, to the best of our knowledge, the idea behind customization of diagnostic tests has not been explored before. Figure \ref{fig:overview} shows the overview of our \textit{personalized diagnosis} system.


Esophageal cancer is cancer arising from the esophagus (8 inches long muscular tube connecting the pharynx with the stomach) usually accompanied by symptoms such pain while swallowing, hoarse voice, etc. It is observed in both developed and developing countries, with  causes including various forms of consumption of tobacco, insufficient intake of fruit and vegetables, overweight and obesity, alcohol consumption, acid reflux disease, etc. The rate of esophageal cancer is rising worldwide, even though the onset and magnitude varies among countries \citep{bibli3}. American Cancer Society estimates approximately $17000$ new cases of esophageal cancer in 2015 \citep{bibli4}. Esophageal cancer rate has also been increased in Great Britain since late 1970s \citep{bibli5}.

A variant of esophageal cancer, called squamous cell carcinoma \citep{Squamous}, appearing mostly in the upper esophagus, is also common in developing countries such as India. Funded by Govt. of India, a reputed hospital in Mumbai, India floated two mobile vans, with the objective of diagnosing esophageal cancer in rural Maharashtra (a state of India). They collected a host of features, ranging from lifestyle, medical history to clinical test features (see section \ref{sec:dataset} for details). We use this dataset, collected for over $20000$ patients to perform the studies in this paper. The broad steps taken for the study are:
\begin{itemize}
\item Collect and preprocess electronic health records (EHR) into features for a machine learning (classification) problem.
\item Design the "best" classifier for the purpose, using all the features.
\item Design a methodology for classification such that no diseased patient get classified as "normal".
\item Remove clinical test features, one by one and find the best classifiers under the above constraint.
\item Assign "costs" to features and find the best set of features corresponding to defined cost budget.
\end{itemize}

First we study the designing of classification methodologies for electronic health records, a well studied problem for many diseases e.g. heart disease \cite{bibli10,bibli11,bibli12,bibli13,bibli14}, breast cancer \citep{bibli16,bibli17,bibli18}, etc. (ref section \ref{sec:literature}). We find that classification esophageal cancer has not been studied. Moreover, there is no consensus among the existing studies regarding the best classification technique. Our first result in this study is that kernel methods with SVM and Logistic Regression perform better that existing popular methods for classification, especially when the number of clinical test features is low.

The second crucial aspect is a careful design of metric for determining the best classifier. This is driven by two peculiarities of the problem: imbalance in the classes and differential importance of the classes. Firstly, there are a lot more "normal" patients than "diseased" patients. Hence, we find that traditional metric of accuracy undiscerning and possibly misleading, since marking most patients as normal will automatically have a high accuracy. Secondly, the cost of classifying a diseased patient as non-diseased (called false normals) is much higher that the cost of classifying a normal patient as diseased, since in the second case the patient has to merely perform more tests, while in the first case, the patient risks making the disease worse. Hence, we use sensitivity (ratio of number of diseased patients detected to total number of diseased patients) as our metric, which is more discerning than accuracy. 

We note that sensitivity of 100\% corresponds to the situation where none of the diseased patients are classified as non-diseased. In many classifiers the ratio of examples predicted as positive or negative can be controlled using a parameter. For example, this can be achieved in SVMs and Logistic regression by changing the prediction thresholds. Moreover, in case of logistic regression, the probability threshold lies within the range $[0,1]$ and hence easier to select. In section \ref{sec:thresholdselection}, we described an algorithm for selection of threshold, which makes false normals zero. Under this criterion, the quality of the classifier is determined by the number of normal patients classified as diseased (false abnormals), which is used as a metric in the subsequent results. To the best of our knowledge, this approach for evaluating EHR classification has not been used before.

Finally, in section \ref{sec:case-study} we describe two case studies which demonstrates the ability of patients to express their preferences for selecting clinical tests.
This choice may be based on financial cost, medical value or simply comfort factor for a patient. We select specific sets features (clinical tests) which satisfy a total budget constraint based on costs specified by the patients or doctor or service provider like insurance companies. These sets of features are used to classify patients, keeping the sensitivity at 100\%  using the algorithm developed above. We use the false abnormal rate to assess the efficiency of the resultant system. We report two case studies, one involving the cost of the of tests and the other using "discomfort" factor of the tests. Our results demonstrate that best set of tests, differ from each other based on the criteria (cost or "discomfort") while maintaining sensitivity of 100\%.

The rest of the document has been structured as follows. Section 2 contains data set preparation. This section also contains how the features are being selected based on need of an individual patients and states about the evaluation criteria of performance of the methods. Section 3 states about the methods for detecting the proposed esophageal cancer and selection of threshold. Section 4 report the results and discussion about classification of EHRs and case-studies on personalized selection of tests. We conclude with our remarks in Section 5 with a summary table.

\subsection{Literature Review}
\label{sec:literature}

Health care captures huge amount of patient specific clinical information e.g. diagnosis, medication, pathological test results and radiological imaging data along with patients' socio demographic characteristics \citep{bibli6}. Although EHR has been heralded for its potential but integrating scattered, heterogeneous data, and varieties of data \citep{bibli7} \citep{bibli8} is still a technical challenge to researchers, who wish to analyze large amounts of patient data. Data mining has helped many researchers to reveal the hidden information using EHR. 

Many researchers have applied supervised machine learning algorithm (a data mining tool) to separate the patient class from the population. For example Alizadehsani \textit{et al.}\citep{bibli10} used data mining technique for diagnosis of coronary artery disease and found Sequential Minimal Optimization (SMO) was the best data mining tool for the kind of data they used for their research among Na\"{i}ve Bayes, SMO, Bagging, and Neural Network. However, Peter \textit{et al.}\citep{bibli11} compared Na\"{i}ve Bayes, K-Nearest Neighbour, Decision Tree and Neural Network for prediction of heart disease and found Na\"{i}ve Bayes was the best among all these methods. Nahar \textit{et al.}\citep{bibli12}tried to detect factors, which contribute to heart disease in males and females using association rule mining like Apriori, Predictive Apriori and Tertius method and concluded that association rule mining-based classifiers helped to identify the key factors behind the disease. Austin \textit{et al.}\citep{bibli13} predicted the probability of the presence of one sub-type of heart failure (heart failure with preserved and reduced ejection fraction) in patients with heart failure using boosting, bootstrap aggregation (bagging), Random Forests, and SVM; and found classical logistic regression was the superior comparing other methods. Shouman \textit{et al.} \citep{bibli14} noticed that hybrid data mining method showed promising result for treatment of heart disease and they proposed a new model to use both single and hybrid data mining methodology to reach conclusion. Wu \textit{et al.}\citep{bibli15} found that logistic regression and boosting were the most efficient method and SVM is the worst, may be due to imbalanced data, in predicting heart failure before more than 6 months before clinical diagnosis.

Data mining techniques have also revealed unknown causes and helped in detecting breast cancer. Alolfe \textit{et al.}\citep{bibli16} used k-NN algorithm from the extracted features to classify if a particular region of interest (ROI) of the digital image, an output of mammogram, is carrying benign or malign masses to determine breast cancer and reported that the algorithm at $k=1$ gave the best result. Abreu \textit{et al.}\citep{bibli17} tried to find out overall survival rate for woman, suffering from breast cancer using 3 ensemble methods (TreeBagger, LPBoost and Subspace) considering 25\% missing data and found Treebagger with 3 neighbor was the best among the above three methods. Jacob and Ramani \citep{bibli18} used the Wisconsin prognostic Breast Cancer data set and compared 20 different classification algorithms to compare the performance of these methods to detect the breast cancer and concluded that Quinlan’s C4.5 algorithm was the best. They \citep{bibli19}also proposed an improved method of detecting lung cancer tumor type based on various properties of protein and reported the benefits of using Bayesian Network learning algorithm in this type of research. 

Data mining techniques have also helped in predicting other type of diseases. Kay \textit{et al.}\citep{bibli20} used Logistic Regression to find out the health related quality of life(HRQoL) for Irritable Bowel Syndrome patients and found that Psychological morbidity, marital status and employment status were associated with HRQoL. Brain \textit{et al.}\citep{bibli21} experimented by Artificial Neural Network, Multilayer Perceptron to conclude that HIV status of a person based could be predicted based on demographic data. Altikardes \textit{et al.}\citep{bibli22} studied many classifiers like Decision trees, naive Bayes, support vector machines, voted perceptron, multi-layer perceptron, logistic regression etc. for classification of Non-Dipper or Dipper Blood Pressure Pattern without Holter Device for the patients suffering with Type 2 Diabetes Mellitus. They noticed that Machine learning was able to predict diurnal blood pressure pattern depending on demographic, clinical and laboratory data. Dheeraj \textit{et al.}\citep{bibli23} found random forest algorithm was most efficient in prediction of factors associating with pressure ulcers among Logistic Regression, Decision Trees, Random Forests and Multivariate Adaptive Regression Splines. 
Wuyang \textit{et al.}\citep{bibli30} experimented to predict the hospitalization due to heart disease with help of SVM, AdaBoost with trees, LR, Na\"{i}ve Bayes Event Model and K-Likelihood Ratio Test(K-LRT). They recommended a novel method of using K-LRT for feature identification during examining a patient and reported AdaBoost as best achiever of highest detection rate at fixed false alarm rate.

The above literature review highlight followings:
\begin{enumerate}

\item Although many literature are available for classifying a patient based on EHR for heart disease, breast cancer and some other diseases but hardly any literature are found for classification of esophageal cancer based on demographic, lifestyle and basic clinical data.\\

\item There is no prior work of personalized diagnostic test selection according to the choice of different stakeholders like doctor, patient, health care service provider or insurance companies etc. \\

\item There is no best data mining method across all types of EHR data. However the most used methods are LR,SVM,RF and NB in the literature review for this study.\\

\item Kernel methods are not very popular among the data mining methods used in disease prediction with help of EHR. \\

\end{enumerate}

\section{Dataset and Model Evaluation}
\label{sec:dataset}

In this section, we describe the data preparation methodology, statistics about the collected datasets (section \ref{sec:datasetprep}), and the metrics we use to evaluate our performance of our model (section \ref{sec:metrics}).

\subsection{Data Set Preparation}
\label{sec:datasetprep}

For this research, we have used the data collected by a reputed hospital in Mumbai, India  using two mobile vehicles from the remote areas of Maharashtra, India. They collected the data in eight different data collection forms, each populated by different health care professionals (Doctors, paramedical personnel, Field operator, laboratory technician) and depending on the choice of patient (control or experiment). These data-sets were joined using patient's unique key to form the final electronic health record (EHR).This EHR had 169 fields out of which 143 fields were retained after de-duplication. Further, these 143 fields were scrutinized manually and fields, which contain information encoded in other fields, or which contained information obviously irrelevant to the problem were removed. For example,we selected `Number of years of tobacco chewing' and discarded `Tobacco chewing (Y/N)'. Also, we removed some of the fields for their large variation like `pin-code', `mobile number', etc. The final EHR had 57 high quality fields enumerated in table 1.
Mobile Vehicles collected basic data for 21,142 patients and selected 3689 patients for potential threat of esophageal cancer using rule driven algorithm to allow these patients for doctor's visit and further clinical tests. As these 3689 patients' records are populated with the label (class deterministic field), we conduct experiments only with these patients' records.

There were a few missing values in these records (1.308\% of all the values) which were populated to value based on rule or average. For example, `sex' field was null for 2 instances but filled with `female' because we get value in `Last Menstrual Period' field for these instances. We imputed missing values for all of the fields using such rules, except `height' and `weight'. For `height' and `weight', we imputed missing values using average values for respective field,a total of 10 instances for each.

Some of the fields were nominal, having multiple values. For example `marital status' (Married, Unmarried, Widow. etc.), `Religion' (Buddhist, Christian, Hindu, Muslim, others), `occupation' (7 values), `Double Count Barium Swallow test' (3 values), etc. We have split each of these nominal fields to multiple binary fields, one for each value of the nominal field. For example, `Marital Status' field is split into `Marital Status-M' for married, `Marital Status-U' for unmarried and so on. Thus we constructed 82 fields(including label) contains value 1 or 0 from original 57 nominal fields and call these organized binary medical fields as features.

\begin{table}[h]
\centering
\begin{tabular}{l l l l }
\hline
\textbf{Category} & \textbf{No. of } & \textbf{No. of }& \textbf{Examples}\\
\textbf{ } & \textbf{Nominal } & \textbf{Binary}& \textbf{ }\\
\textbf{ } & \textbf{Features } & \textbf{Features}& \textbf{ }\\
\hline
Demographics & 12 & 27 & Sex, Age, \\
&&&weight,height etc.\\
\hline
Lifestyle & 17 & 17 & Tea consumption , \\
&&& Duration of cigarette smoking, \\
&&& Duration of tobacco chewing, etc.\\
\hline
Basic Clinical Tests  & 15 & 15 & Diabetes, Blood Pressure, \\
(BCT)&&& Asthma, Tuberculosis, \\
&&& cardiac disease etc.\\
\hline
Medical Practitioner  & 8 & 18 & Erythroplakia,  \\ 
(MP)&&& ErythroLeukoplakia,\\
&&& MucousFibrosis, ulcer etc.\\
\hline
History & 4 & 4 & Cancer case in last \\
&&& 5 years in family, \\
&&&Any cancer death in family\\
\hline
Label & 1 & 1 & Oral Cavity.\\
\hline
\end{tabular}
\caption{Category Wise Various Fields}\label{table:features}
\end{table}

The features have been shown in the Table 1 according to various categories along with some examples in each category. We define our problem as a classification problem assigning each patient (as indicated in label) either `Abnormal', if there is an esophageal-cancer suspect and `Normal' otherwise.

\subsection{Model Evaluation Criteria}
\label{sec:metrics}

In medical diagnosis, an outcome of a test is called negative if a person is not detected with the disease (esophageal cancer here).In the data-set, a person diagnosed with esophageal cancer is described as `abnormal', `normal' otherwise.Hence, we use `abnormal' or `positive' and `normal' or `negative' interchangeably.The confusion matrix is given as below.\\
\[
\left(
\begin{array}{cc}
TA&FA \\
FN&TN
\end{array}
\right)
\]

where, TN: True Normal(where the model predicts an esophageal cancer negative person as normal), FA(or FP): False Abnormal(where model predicts an esophageal cancer negative person as abnormal), FN: False Normal(where model predicts an esophageal cancer positive person as normal) and TA(or TP): True Abnormal(where model predicts an esophageal cancer positive person as abnormal).

The biggest issue with automatic diagnosis is classification of a diseased person (positive) as normal (negative), as this can give a false sense of security to the patient resulting in deterioration of his condition. Hence, we first report the sensitivity \citep{bibli31}of our classifiers, which is defined as the fraction of abnormal (positive) patients as positive.

\begin{equation}
\label{sensitivity}
sensitivity =  \frac{TA}{TA+FN}
\end{equation}
Also, we analyzed the result with help of two metrics(FN and FA), mentioned above \citep{bibli32}. We also used the standard definition of accuracy\citep{bibli33}.
\begin{equation}
accuracy = \frac{(TA+TN)}{(TA+FA+TN+FN)} 
\end{equation}

We have used 66.67\%-33.33\% split for training and performance evaluation respectively for the four methods chosen in this research using both the data-set with 82 features and 64 feature set respectively. The same testing scheme applies to kernel method after feature transformation to a higher dimension for SVM and LR as described in the next section.

\section{Methodology}
\label{sec:methodology}

In this section, we describe the Machine Learning techniques which are commonly used for automatic classification of patients using EHRs (section \ref{sec:methods}). While the methods performed very well with all features, removal of features caused a degradation in performance of linear logistic regression. Hence, we used kernel logistic regression \cite{zhu2001kernel}, which is described in section \ref{sec:kernel}. Finally, in section \ref{sec:thresholdselection}, we describe our method of selecting probability threshold for classification with logistic regression which ensures zero false negative rates which is critical for this application.

\subsection{Methods Description}
\label{sec:methods}

The problem of predicting whether a patient is a suspect of esophageal cancer or not, using available medical history through EHR data, is naturally posed as a binary classification problem. This has been studied extensively in the Machine Learning literature. Four methods stand out as the most popular choices: Logistic regression (LR) \cite{bibli29}, Support Vector Machines (SVM) \cite{bibli27,bibli28}, Naive Bayes (NB) \cite{bibli24}, and Random Forests (RF) \cite{bibli25, bibli26}. These have also been used for classification of EHRs for other diseases like lung cancer \cite{bibli18}, etc. We conduct experiments using these four methods and report the results in section \ref{sec:results}. In this section, we describe each of these techniques briefly, and also describe characteristics of these techniques which make them suitable for the application.

\subsubsection{Na\"{i}ve Bayes} 
\label{sec:nb}

Na\"{i}ve Bayes classifiers are a family of simple probabilistic classifiers based on applying Bayes' theorem with Na\"{i}ve (strong) independence assumptions between the features conditioned on the class label, called class conditional independence \cite{bibli24}. The assumption simplifies the computation, in the sense that it has less parameters to compute.
The posterior probability is calculated based on likelihood, class prior probability and predictor prior probability and due the Na\"{i}ve assumption $P(C|X)$ has also been simplified as stated below.
\begin{equation}
\label{NB}
P(c|x)= \frac{P(x|c)P(c)}{P(x)}
\end{equation}
where $P(c|x)$ is Posterior Probability, P(x) is  Predictor Prior probability, P(c) is Class Prior Probability and $P(x|c)$ is called Likelihood.


Despite the stated simplicity and assumptions, Na\"{i}ve Bayes classifiers work well in many practical situations. It can model non-linear class boundaries and is theoretically optimal in a certain sense \cite{bibli24}. For this paper we use the Naive Bayes Classifier implemented in open source software package, Weka \cite{weka}.  

\subsubsection{Random Forest}
\label{sec:RF}

A random forest,introduced by Leo Breiman \cite{bibli25}, is a classifier based on ensemble methods that engages a number of decision tree classifiers on various sub-samples of the data-set. 
In decision trees, each node is split using the best split among all variables but in random forest each node is split using the best among a subset of predictors, chosen randomly at that node. Hence, feature selection is not greedy, but multiple such trees are constructed to boost the accuracy. Moreover, the depth of each tree is limited, thus prevent overfitting.

Random forests are popular tools for modelling complex relationships between features. However, they are computationally very expensive, due the need for training multiple trees. We used the implementation of random forests provided in Weka \cite{weka}. They have two tunable parameters: the number of variables in the random subset at each node and the number of trees in the forest. We found that the classifier  is not very sensitive to parameter values, and is very user friendly in nature and the procedure is consistent and adapts to sparsity, as its rate of convergence depends only on the number of strong features and not on number of noisy features \citep{bibli26}.

\subsubsection{Support Vector Machine}
\label{sec:SVM}

Support Vector Machine is a supervised learning model, which constructs a hyper-plane or set of hyper-planes in the feature space (in a high- or infinite-dimensional) that can separate the data points belong to different classes residing the data points on different sides of that hyper plane. The minimum over all the distances of each data point from the hyper-plane is called margin. Intuitively a good separation is achieved when the margin is largest, as the generalization error of the classifier becomes lower when the margin becomes larger \citep{bibli27}.

When the data points are not linearly separable, the classifier can be made adaptive to some mis-classification errors and elevate the features into a higher dimensional space to make the data points linearly separable. We employed the widely used Radial Basis Function (RBF) and polynomial as the kernel function in our experiment settings \citep{bibli28}. 

\subsubsection{Logistic regression}
\label{sec:LR}

Logistic regression is a classification method by which an example / datapoint can be categorized into any one of the two mutually exclusive and exhaustive classes.  The method models the posterior probability that a sample being classified in the positive class (e.g., the True class) as a logistic function of linear combination of input features:
\begin{equation}
P(x_i) = P(y_i = 1 | w, x_i) = \frac{1}{1+e^{w^T x_i}}
\end{equation}
Here, $w$ is the estimated parameter.
Hence the decision boundary of the two class separator is linear.
We have used WEKA tool for initial analysis which uses LR algorithm according to LeCessie and van Houwelingen \citep{bibli29} and the (negative) binomial log-likelihood form for which $L(w)$ is minimized is thus:

\begin{eqnarray}
L(w) = -\sum_{i=1}^{n}[(y_i*ln(P(x_i)) + (1 - y_i)*ln(1-P(x_i))] + ridge*(\|w\|^2)
\end{eqnarray}

In the above equation L2 regularization has been used. We also tried using the L1-regularized logistic regression algorithm supported by the Liblinear tool \cite{liblin}. The probability of positive class is same as above while the objective function is given by:

\begin{equation}
L(w) = C(\sum_{i=1}^{l} log(1+e^{-w^Tx_i)}+\sum_{i:y_i=-1}w^Tx_i) + \lambda \|w\|_1
\end{equation}

In both the above problems, the final solution is given by $w^* = \mbox{arg}\min_w L(w)$. While this solution defines a linear boundary between the positive and the negative classes, one can use feature expansion or kernel trick to learn non-linear boundaries. In the next section, we describe these methods. As we shall see in section \ref{sec:results-kernels}, L1 regularization along with non-linear feature map yields the best results for our problem.


\subsection{Kernel Methods for Patient Classification}
\label{sec:kernel}

The four popular methods, as described in section \ref{sec:methods} did not perform well on removal of MP features from the dataset as shown in section \ref{sec:results-allfeatures}, i.e. only demographics, lifestyle, basic medical tests, and medical history based features are not able to predict an esophageal cancer patient with a good accuracy. However, on manual inspection it was found that these features have the necessary information for classification of esophageal cancer. Also, unlike the case where medical practitioner features were included, both Naive-Bayes and Random Forests give better results than linear-SVM and LR. This information hints towards a non-linear decision boundary between the classes.

Kernel methods were introduced by Smola and Scholkopf \cite{smolabook} as a "trick" for generating non-linear classifiers with linear methods like SVM and LR. The key tricks involve (1) expressing the training  problem as well as the classification function as a one which only relies on the dot product between examples, and (2) defining new kernel functions, which emulate these dot products in very high dimensional spaces, the projection of which onto the example space is a non-linear boundary. For both SVMs and logistic regression, the classification function is of the form:
\begin{align}
f(x) &= sign ( \sum_{i=1}^{N} \alpha_i K(x_i,x) ) \label{eq:classificationfunction}
\end{align}
where $K$ is the kernel function and $\alpha_i$ are the Lagrange multipliers corresponding to training of dual problems. $sign$ function is defined as $sign(x) = +1$ if $x\geq 0$, $-1$ otherwise.\\
For SVMs, the dual training problem becomes:
\begin{align}
& \max_{\bm{\alpha}}\mathcal{L}(\vec{\alpha}) = \sum_{i}\alpha_{i} - \frac{1}{2} \sum_{i,j}\alpha_{i}\alpha_{j} y_{i}y_{j} K(x_i,x_j) \label{eq:SVMdual} \\
& \mbox{\textbf{subject to:}  }  0 \leq \alpha_{i} \leq C\ \ \ \forall i \in {1,\cdots, N}  \nonumber \\
&  \sum_{i=1}^{N} \alpha_i y_i = 0  \nonumber
\end{align}
where $(x_i,y_i),\ i=1,\cdots,N$ are the training data points, and $\alpha_i$ are the Lagrange multipliers (dual variables). $C$ is a user supplied constant that governs the trade-off between accuracy and generalization.

For Logistic Regression, the dual training problem was proposed in \cite{zhu2001kernel}, as:\\
\begin{equation}
\min_{\vec{a}} H(\vec{a}) = -\vec{y}^T(K_a\vec{a})+\vec{1}^T ln(1+exp(K_a\vec{a}))+\frac{\lambda}{2}\vec{a}^TK_a\vec{a}
\end{equation}
where $\vec{a} = (a_1,...a_N)^T$; the regressor matrix $K_a = [K(x_i,x_j)]_{N \times N}$. However, we could not find any open source solver for the above problem. Hence, we used the primal formulation along with polynomial feature expansion as described in section \ref{sec:results-kernels}. 

\subsection{Threshold Selection for Logistic Regression}
\label{sec:thresholdselection}

In the previous section, we have described non-linear classification using SVM and LR. As can be seen from section \ref{sec:results-kernels}, kernelized versions of both SVM and LR yield good accuracy, even in the absence of medical-practitioner features. Normally, one would use equation \ref{eq:classificationfunction} to classify an existing example. This uses a threshold of $0$ on linear map $w^T x$ as the decision boundary, which corresponding to a probability threshold of $0.5$ in case of logistic regression. This is based on the assumption that both classes are equally likely and also equally important.
However, as was discussed before, in our case classification of a patient as a non patient is much more expensive, as it it may cause major damage to the person's health, compared to the scenario where a non-patient is classified as patient, which will only cause inconvenience.

From the above discussion, it is clear that this stucy presents a situation of imbalanced classification, which has been studied in the machine learning literature in many other contexts \cite{chawla2002smote}. Most techniques for imbalanced classification focus on incorporating class sensitive penalties at the time of training. One way to do this is to re-sample the training set, so as to obtain a specified ratio of positive and negative samples. We use technique to improve the training accuracy, described in section \ref{sec:results-kernels} as "un-skewing". However, this technique does not guarantee a zero mis-classification from abnormal to normal class.

In this article, we propose to change the threshold for the classification function based on misclassifications in the test dataset, to attain zero misclassification error from abnormal to normal (patient to non-patient) classes. This corresponds to achieving a sensitivity of 1 (one). The objective here is to come up with an algorithm which can figure out a suitable threshold. This is particularly tricky since the space for threshold in the linear map space ($w^Tx$) is unbounded. Here, we propose to alter the threshold in the probability space ($\sigma(w^Tx)$) which is bounded between $0$ -- $1$. However, despite several attempts, there is no consistent way of converting the SVM "margin" into probability scores. Hence we stick to LR for the current task. Algorithm \ref{alg:thresholdselect} describes the algorithm for selecting threshold, making false-normal zero. In the next section, we report results for various experiments performed to validate the ideas proposed here.


\begin{algorithm}[h]
 \KwData{Probability output for test instances, $z_1,\dots,z_n$; 
  Actual class values (normal / abnormal);}
 \textbf {Initialization:} $Threshold$ to 0.5; $False Abnormal$ to 0\;
   \tcc{Calculate Threshold}
   \For{$i=1$ to $n$}
    {
      \If{$z_i > Threshold$}
       {
        \If{actuallabel($i$) $\neq$  normal}
          {$Threshold = z_i$}
        } 
    }
    \tcc{Calculate False Abnormal}
   \For{$i=1$ to $n$}
    {
      \If{ ($z_i < Threshold$) and (actuallabel($i$) $=$  normal)}
      { Increment $FA$ by $1$ } 
    }
    return $FA$ as False Abnormal
\caption{False Abnormal value to make False Normal zero} \label{alg:thresholdselect}
\end{algorithm}

\section{Results and Discussion}
\label{sec:results}

The methods developed in the section \ref{sec:methodology} are implemented and tested on the dataset described in section \ref{sec:dataset}. Although all four methods are initially tested using Weka \cite{weka}, but later Logistic regression (LR) is implemented using liblinear \cite{liblin} for scalability. For kernelized SVM we used Libsvm \cite{libsvm} from WEKA. Code for generation of features and searching for appropriate threshold was written in Python.

Section \ref{sec:results-allfeatures} reports the prediction accuracy using all the features (corresponding to various tests) and also without using features that involve trained medical practitioners, using the techniques described in section \ref{sec:methods}. Section \ref{sec:results-kernels} reports results of automatic classification.

\subsection{Prediction with and without MP features}
\label{sec:results-allfeatures}

Initially, we evaluate the performance of  the four methods, described above using 82 features in terms of accuracy and sensitivity. A comparison of the accuracy and sensitivity of these methods has been reported in the table \ref{table:sensitivity-all-features} and table \ref{table:accuracy-all-features}.
\begin{table}[h]
\centering
\begin{tabular}{l| l l l l}
\hline
\textbf{Data-Set } & \textbf{LR} & \textbf{SMO} & \textbf{RF}& \textbf{NB}\\
\hline
\multicolumn{5}{l}{\textbf{With MP Features}} \\
\hline
Training & 100 & 97.83 & 100& 97.83\\
Test & 93.75 & 93.75& 93.75& 81.25\\
\hline
\multicolumn{5}{l}{\textbf{Without MP Features}} \\
\hline
Training & 9.78  & 8.70& 47.83& 34.78 \\
Test & 12.50  & 12.50 & 12.50& 12.50  \\
\hline
\end{tabular}
\caption{Sensitivity in \% with all features} \label{table:sensitivity-all-features}
\end{table}

\begin{table}[h]
\centering
\begin{tabular}{l| l l l l}
\hline
\textbf{Data-Set } & \textbf{LR} & \textbf{SMO} & \textbf{RF}& \textbf{NB}\\
\hline
\multicolumn{5}{l}{\textbf{With MP Features}} \\
\hline
Training & 100 & 99.95& 100& 99.26\\
Test & 99.84 & 99.84& 99.84& 99.36\\
\hline
\multicolumn{5}{l}{\textbf{Without MP Features}} \\
\hline
Training & 97.64  & 97.64& 98.67& 88.37 \\
Test & 97.29  &97.45 &97.53& 97.29 \\
\hline
\end{tabular}
\caption{Accuracy in \% with all features} \label{table:accuracy-all-features}
\end{table}

The result appears promising when we conduct this experiment with all 82 features i.e. inclusive of features measured with help of medical practitioner (MP). However, the experiment after removing the $18$ MP features shows none of the classifiers is able to classify the patients from the population. This indicates that without medical practitioner's supervision these methods are not capable to predict esophageal cancer with help of only demographic, lifestyle, basic clinical test and patient history data.

\subsection{Prediction accuracy with Kernels}
\label{sec:results-kernels}

We observe that all the four methods are not able to achieve the desired sensitivity but still maintain a high accuracy due to skewness of the data. We "unskew" \cite{chawla2002smote} the data set by supersampling the abnormal class $10$ times, reaching at a total instances of 4609, with a class ratio (Abnormal to normal) of $0.2813$.  We perform the experiments with the four methods again but are not able to attain zero false normal rate (see table \ref{table:sensitivity-unskewed-withoutMP}).

Although the above four methods do not yield desired result but we noticed that the data has the potential for the prediction with a high accuracy. For example, `weight loss' feature at value 1 has only abnormal class. Similarly `Tuberculosis', `Asthma', `Alteration of Voice', `Neck-nodes' etc. have only values at a class 0 or 1. Also, we find that RF and NB gives a sensitivity much higher than LR and SMO (table \ref {table:sensitivity-unskewed-withoutMP}), which directs us towards a non linear decision boundary for this classification. We decided to transform the data-set to a higher dimension with the unskewed data. 

In table \ref {table:sensitivity-unskewed-withoutMP}, we report the results for fitting high dimensional models using kernel methods. First, we test with polynomial kernel of degree 2
 with SVM using LIBSVM tool within WEKA, which does not yield a sensitivity $100\%$. However, a kernel of degree 3 results in $100\%$ sensitivity. This supports our earlier findings about the potential of the data-set for giving a prediction at a higher dimension.

\begin{figure}[h]
\centering\includegraphics[width=.8\linewidth]{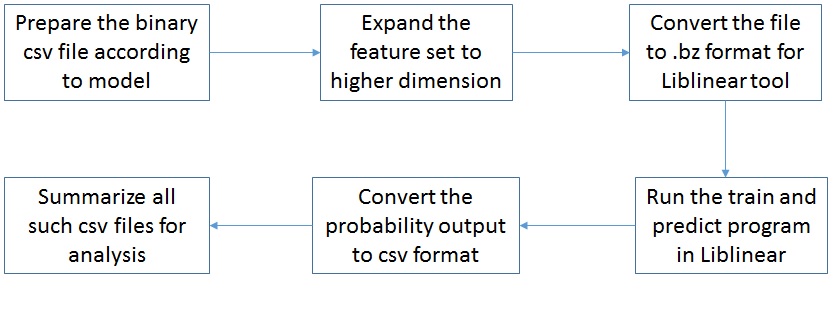}
\caption{Operational Flowchart}
\label{fig:operational-flowchart}
\end{figure}

Since, we are interested in the subsequent threshold selection as described in section \ref{sec:methodology}, we need to use LR. However, none of the popular ML package packages (including WEKA and libsvm) support kernel LR. So, we generate degree 3 polynomial kernels for LR with python program.
We expand $63$ features in the input space (except label) with bias 1, 
to $45760$ features in the higher dimensional space consisting of all monomials of degree upto 3.
We then experimented with linear LR using Liblinear tool, for classification of patients.  The process flow diagram is shown in fig \ref{fig:operational-flowchart}.

Finally we observe that for kernel of degree 3, LR achieves 100\% sensitivity (table \ref{table:sensitivity-unskewed-withoutMP}) as expected.

\begin{table}[!h]
\centering
\resizebox{\linewidth}{!}{%
\begin{tabular}{l| l l l l l l l}
\hline
\textbf{Data-Set } & \textbf{LR} & \textbf{SMO} & \textbf{RF}& \textbf{NB} & \textbf{Kernel}& \textbf{Kernel}& \textbf{Kernel}\\
\textbf{ } & \textbf{} & \textbf{} & \textbf{}& \textbf{}& \textbf{SVM}& \textbf{SVM} & \textbf{LR}\\
\textbf{ } & \textbf{} & \textbf{} & \textbf{}& \textbf{}& \textbf{(Degree 2)}& \textbf{(Degree 3)} & \textbf{(Degree 3)}\\
\hline
& \textbf{Sensitivity}\\
\hline
Training &  33.70  & 28.09& 53.26 & 58.70 & 90.22& 100.00 & 100.00\\
Test & 32.42  & 24.45& 48.35& 49.18& 83.79  & 100.00  & 100.00\\
\hline
& \textbf{Accuracy}\\
\hline
Training &  82.47  & 81.06 & 89.74 & 72.99 & 95.96 & 100.00  & 99.80\\
Test & 81.37  & 79.90 & 87.94 & 75.56 & 92.41 & 94.89 & 91.08\\
\end{tabular}}
\caption{Sensitivity and Accuracy in \% for unskewed data without MP features}\label{table:sensitivity-unskewed-withoutMP}
\end{table}


\subsection{Trade-off between FA and FN using Threshold}
\label{sec:results-threshold}

In the previous section, we saw that it is possible to classify the patients as normal or abnormal, without using only features derived from tests conducted by Medical practitioners, and ensuring that false normal rate is zero. However, the classifier still uses $15$ features from "basic clinical tests" category, which also may be inconvenient or expensive to acquire. In this section, we attempt to remove the these features from the classification.

While reducing the "basic clinical test" (BCT) features (table \ref{table:features}) from 63 (excluding label), one by one 
upto 49 features, we got zero false normal using a classification threshold of $0.5$ (default value) till 59 features. On further de-selection of features, we had to increase the the threshold to get zero false normal. We used the algorithm described in section \ref{sec:thresholdselection} selection of threshold.
For all experiments in this section and the next section, we use kernel logistic regression, along with this threshold selection algorithm.

\begin{figure}[h]
\centering\includegraphics[width=.8\linewidth]{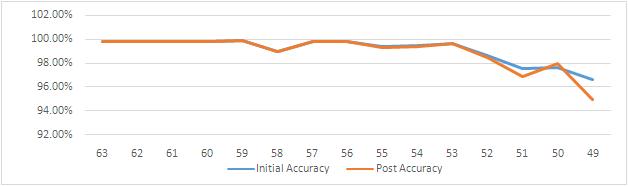}
\caption{Initial Accuracy \& Post Accuracy (after making $FN=0$) for various number of Features}
\label{fig:initaccuracy}
\end{figure}

Figure \ref{fig:initaccuracy} shows the accuracies obtained by predicting using threshold of $0.5$, called initial accuracy (blue line), and threshold obtained for zero false normal, called post accuracy (brown line), for various number of features used. As expected, the accuracies decrease as the number of features decreases (though it remains constant till 53 features). We also see that the accuracies obtained using the threshold for zero false normal rate are lower than those using the threshold of $0.5$. The drop can be attributed to rise in false abnormal rate.


\begin{figure}[!h]
\centering\includegraphics[width=.8\linewidth]{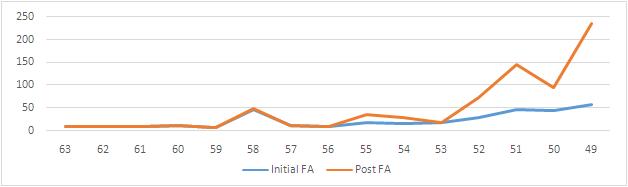}
\caption{Initial FA \& Post FA (after making $FN =0$) for various number of Features}
\label{fig:initFP}
\end{figure}

The increase in false abnormal rate is reflected in the number of normal patients being diagnosed as abnormal and asked to go through additional diagnosis. In Figure \ref{fig:initFP}, we report the number of such patients out of a total 4609 patients, for threshold of $0.5$ (blue line) and threshold for zero false normal (brown line). We observe that the additional number of patients sent for further diagnosis is $\sim 200$ even after removing all test features, i.e. using only demographics, lifestyle and medical history based features.
These charts can also potentially be used in determining the optimal number features considering the opportunity cost for the change in FA or accuracy with respect to number of features. In the next section, we describe case studies selection of features when various objectives corresponding to cost and comfort for tests are considered.

\subsection{Case studies: Test Selection using Cost and Comfort}
\label{sec:case-study}

The algorithm designed in section \ref{sec:thresholdselection} can be used for personalized selection of test schedule for individual patients. In this section, we present two such case studies based on two common concerns of patients while undergoing medical diagnosis: \textbf{cost} and \textbf{comfort}.
For the case study involving cost of tests, the prices have been taken in INR from sources: \citep{bibli34}\citep{bibli35}. It is understood that the prices are only indicative, and will vary across diagnostic centers. Also, for the study only relative prices are important. For the case study involving comfort preferences, in absence of standard measures, we consider some natural values for \textit{discomfort index} (values between 1 to 10). The values for different test are are listed in table \ref{table:costcomfort}.


\begin{table}[!h]
\centering
\begin{tabular}{l l l}
\hline
\textbf{ } & \textbf{Cost} & \textbf{Discomfort} \\
\hline
\textbf{Systolic } & \textbf{0 } & \textbf{0.5}\\
\textbf{Diastolic} & \textbf{0 } & \textbf{0.5}\\
\textbf{RBSL By Glucometer } & \textbf{50 } & \textbf{1}\\
\textbf{OralHygiene } & \textbf{0 } & \textbf{0}\\
\textbf{Neck Nodes } & \textbf{300 } & \textbf{3}\\
\textbf{Diabetes } & \textbf{500 } & \textbf{5}\\
\textbf{Hypertension} & \textbf{100 } & \textbf{2}\\
\textbf{Cardiac Disease} & \textbf{2000} & \textbf{10}\\
\textbf{Tuberculosis } & \textbf{500 } & \textbf{5}\\
\textbf{Asthma } & \textbf{1500 } & \textbf{10}\\
\textbf{Difficulty in Swallowing} & \textbf{0 } & \textbf{0}\\
\textbf{Alteration Of Voices } & \textbf{0 } & \textbf{0}\\
\textbf{Reflux Gastritis  } & \textbf{1000 } & \textbf{10}\\
\textbf{Haematemesis } & \textbf{300 } & \textbf{3}\\
\textbf{Weight Loss } & \textbf{0 } & \textbf{0}\\
\hline
\end{tabular}
\caption{Cost and Discomfort index indicative values}\label{table:costcomfort}
\end{table}

For the case study, we assume the setting where a \textit{budget} is imposed on the total cost or discomfort during tests performed for diagnosis. Such a constraint can be imposed due to economic condition of a patient, or as decided by insurance provider, or due to a limit on total budget (as set by world bank, government health organization etc.) for a medical project.
The obvious objective is to minimize the false abnormal rate, under the constraints that false normal is zero and the total cost of tests is less than or equal to the budget. 
However, it is easy to see that this problem is NP-Hard, since it involves enumerating all test combinations which satisfy the constraint (assuming the false normal can always be made zero using algorithm \ref{alg:thresholdselect}), and evaluating the false abnormal rate for these combinations. While approaches for tacking this problem will be studied in another work, in this work we follow the following heuristic scheme:
\begin{enumerate}
 \item Set budget on total cost or discomfort value of patient as constraint.
 \item Enumerate all possible options  (combinations of tests) which maximally satisfy the constraint, i.e. for every option non more tests can be added to set, without violating the constraint.
 \item Prepare the feature set by either selecting the tests (cost constraint) or deselecting the tests (choice constraints).
 \item For each option adjust threshold to make $FN = 0$ using algorithm \ref{alg:thresholdselect}; Calculate $FA$.
 \item Choose the option corresponding to minimum $FA$ (False abnormal rate).
\end{enumerate} 

\begin{figure}[!h]
\centering\includegraphics[width=.9\linewidth]{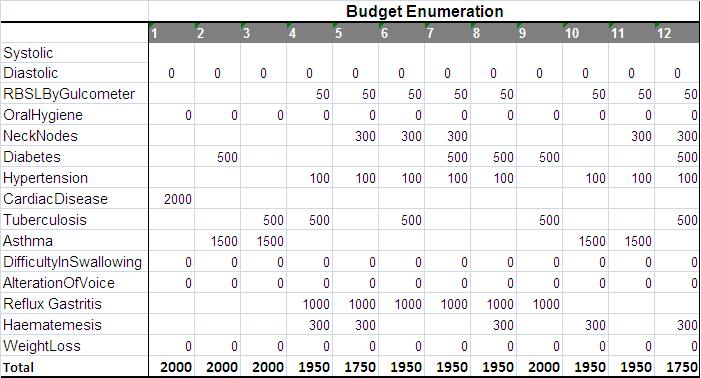}
\caption{Total Possible Enumeration for a Budget Constraint INR 2000}\label{fig:enubudget}
\end{figure}
\begin{figure}[!h]
\centering\includegraphics[width=.9\linewidth]{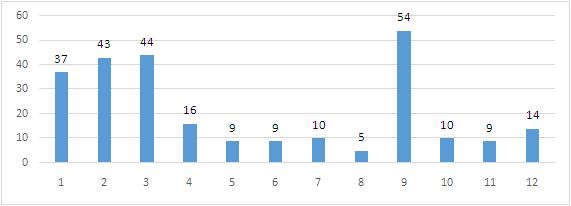}
\caption {Number of False Abnormal for various options of budget}\label{fig:FPbudget}
\end{figure}
For the case study related to costs, we assume a budget of INR 2000 (against  a total cost of all 15 tests INR 6250). Figure \ref{fig:enubudget} enumerates a total 12 options of selecting tests, and figure \ref{fig:FPbudget} shows the corresponding false abnormals out of total 4609 patients. In this case option $8$ is produces minimum number of false abnormals for budget constraint, thus becoming the best option in this enumeration list. This information helps all stakeholder to perform the particular group of tests without going for all the clinical tests with such budget constraint.

\begin{figure}[!h]
\centering\includegraphics[width=.9\linewidth]{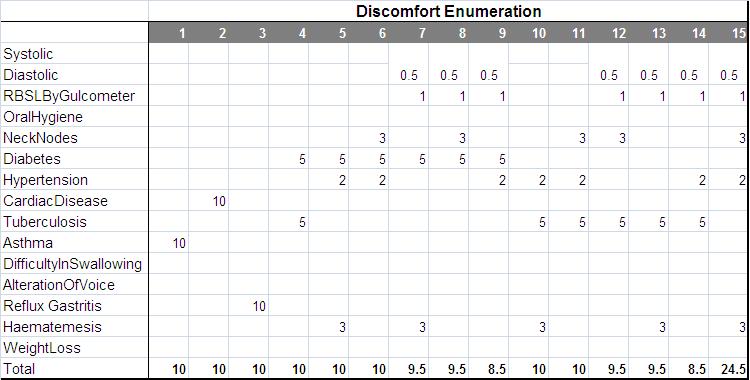}
\caption{Total Possible Enumeration for a Discomfort Value 10}\label{fig:enuchoice}
\end{figure}
\begin{figure}[!h]
\centering\includegraphics[width=.9\linewidth]{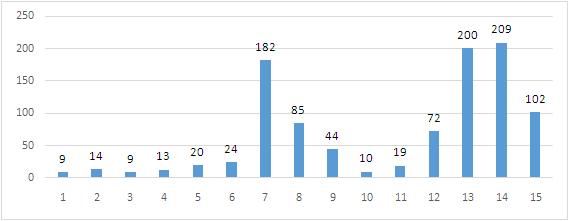}
\caption{Number of False Abnormal for various options of discomfort}\label{fig:FPchoice}
\end{figure}

Similarly, for the case study involving budget on total patient discomfort, we choose a total budget of $10$, which is an upper bound on total discomfort values to be removed. Figure \ref{fig:enuchoice} shows 15 options for selecting tests, and figure \ref{fig:FPchoice} shows the  false abnormals obtained for each choice out of a total of  4609 patients. In our case, option 2 yields the minimum number of false abnormals. This technique can be used to provide patients with a choice to prioritize his/her own tests.

From both the case studies, it is clear that the technique proposed here produces significant benefits over arbitrary test selection practiced today. As can be seen the false abnormal rates vary widely between different options of test selections. To the best of our knowledge, this is the first study of automatic diagnosis system which focuses on making false normals zero, and studying the consequent false abnormal rates.

\pagebreak

\section{Conclusions}
Research on the predictions of diseases using EHR with aid of machine learning techniques has been prevalent in the contemporary literature.
This study illustrates that a machine learning based algorithm can facilitate in the prediction of an esophageal cancer relying on demographic, lifestyle, personal history and customized clinical data (without help of a doctor’s intervention). The novel concept of selecting and de-selecting of clinical tests brings a new dimension to all stakeholders in health industry towards optimization of cost and freedom of choice to have the clinical tests without compromising the detection of all true patients.  We consider that the outcome of this study is computational innovation and societal advancement, the former unveiling a new prediction technique for classifying esophageal cancer patients with a very high accuracy upto $99.80\%$ with a sensitivity $100\%$, and the latter allowing to choose the clinical tests as per wish of either patient, doctor or service provider. Future research directions to this study include application of this computational technique in diagnosis of other diseases, and 
development of algorithms for selecting the optimal feature set as per all choice that could be applied to any disease.\\

\begin{framed}

\large \textbf {Summary Table}\\

\normalsize \textbf {Known artifacts before this study}\\
\begin{itemize}

\item EHR helps in predicting cardiac problems, breast cancer, lung tumor and many other diseases but yet to contribute in predicting esophageal cancer, which is rising worldwide.
 
\item Machine learning plays as a critical instrument in early detection of a disease. 
\end{itemize}

\textbf {Knowledge addition by this research}\\

\begin{itemize}
\item Demographic, lifestyle and basic clinical data (without doctor's supervision) can predict Esophageal cancer with a very high accuracy.

\item A feature transformation to a higher space can have a higher accuracy than orthodox machine learning methods and yield a high sensitivity to detect all true patients.

\item Customized tests using a subset of features(tests) still can predict the esophageal cancer without compromising the probability of non inclusion of a true patient. 

\item The novel idea of selecting a subset of standard tests to predict a disease can help many stakeholders - patient, doctor, insurance provider and others by reducing cost, improving quality or optimizing service parameters.

\end{itemize}
\end{framed}

\pagebreak




\begin{thebibliography}{10}
\expandafter\ifx\csname url\endcsname\relax
  \def\url#1{\texttt{#1}}\fi
\expandafter\ifx\csname urlprefix\endcsname\relax\def\urlprefix{URL }\fi
\expandafter\ifx\csname href\endcsname\relax
  \def\href#1#2{#2} \def\path#1{#1}\fi

\bibitem{obama}
PMC-Corporate,
  \href{http://www.personalizedmedicinecoalition.org/Userfiles/PMC-Corporate/file/PMC_2015_annual_report.pdf}{Pmc
  2015 annual report}, [Online; accessed 21-Dec-2015].
\newline\urlprefix\url{http://www.personalizedmedicinecoalition.org/Userfiles/PMC-Corporate/file/PMC_2015_annual_report.pdf}

\bibitem{bibli3}
E.~G., A.~HO, N.~O. Weiderpass~E, A global assessment of the oesophageal
  adenocarcinoma epidemic, An International Journal of Gastroenterology and
  Hepatology\href {http://dx.doi.org/10.1136/gutjnl-2012-302412}
  {\path{doi:10.1136/gutjnl-2012-302412}}.

\bibitem{bibli4}
B.~Scott, W.~Health,
  \href{http://www.news-medical.net/news/20150421/Incidence-of-esophageal-cancer-linked-to-GERD-rises-six-fold
  -in-recent-decades.aspx}{Incidence of esophageal cancer linked to gerd},
  [Online; accessed 21-Dec-2015].
\newline\urlprefix\url{http://www.news-medical.net/news/20150421/Incidence-of-esophageal-cancer-linked-to-GERD-rises-six-fold
  -in-recent-decades.aspx}

\bibitem{bibli5}
Cancer-Research-UK,
  \href{http://www.cancerresearchuk.org/content/oesophageal-cancer-incidence-statistics#ref-2}{Oesophageal
  cancer incidence statistics}, [Online; accessed 21-Dec-2015].
\newline\urlprefix\url{http://www.cancerresearchuk.org/content/oesophageal-cancer-incidence-statistics#ref-2}

\bibitem{Squamous}
W.~Blot, J.~McLaughlin, \href{http://europepmc.org/abstract/MED/10566604}{The
  changing epidemiology of esophageal cancer}, Seminars in oncology 26~(5 Suppl
  15) (1999) 2â€”8.
\newline\urlprefix\url{http://europepmc.org/abstract/MED/10566604}

\bibitem{bibli10}
R.~Alizadehsani, J.~Habibi, M.~J. Hosseini, H.~Mashayekhi, R.~Boghrati,
  A.~Ghandeharioun, B.~Bahadorian, Z.~A. Sani, A data mining approach for
  diagnosis of coronary artery disease, Computer Methods and Programs in
  Biomedicine (111)~(1) (2013) 52 -- 61.
\newblock \href
  {http://dx.doi.org/http://dx.doi.org/10.1016/j.cmpb.2013.03.004}
  {\path{doi:http://dx.doi.org/10.1016/j.cmpb.2013.03.004}}.

\bibitem{bibli11}
T.~J. Peter, K.~Somasundaram, An empirical study on prediction of heart disease
  using classification data mining techniques, in: Advances in Engineering,
  Science and Management (ICAESM), 2012 International Conference on, 2012, pp.
  514--518.

\bibitem{bibli12}
J.~Nahar, Imam, Tasadduq, Tickle, K.~S., Chen, Y.-P. Phoebe, Association rule
  mining to detect factors which contribute to heart disease in males and
  females, Expert Syst. Appl. (40)~(4) (2013) 1086--1093.
\newblock \href {http://dx.doi.org/10.1016/j.eswa.2012.08.028}
  {\path{doi:10.1016/j.eswa.2012.08.028}}.

\bibitem{bibli13}
P.~C. Austin, J.~V. Tu, Ho, J.~E, Levy, Daniel, Lee, D.~S, Using methods from
  the data-mining and machine-learning literature for disease classification
  and prediction: a case study examining classification of heart failure
  subtypes, Journal of Clinical Epidemiology (66).
\newblock \href {http://dx.doi.org/10.1016/j.jclinepi.2012.11.008}
  {\path{doi:10.1016/j.jclinepi.2012.11.008}}.

\bibitem{bibli14}
M.~Shouman, T.~Turner, R.~Stocker, Using data mining techniques in heart
  disease diagnosis and treatment, in: Electronics, Communications and
  Computers (JEC-ECC), 2012 Japan-Egypt Conference on, 2012, pp. 173--177.
\newblock \href {http://dx.doi.org/10.1109/JEC-ECC.2012.6186978}
  {\path{doi:10.1109/JEC-ECC.2012.6186978}}.

\bibitem{bibli16}
M.~A. Alolfe, A.~B.~M. Youssef, Y.~M. Kadah, A.~S. Mohamed, Development of a
  computer-aided classification system for cancer detection from digital
  mammograms, in: Radio Science Conference, 2008. NRSC 2008. National, 2008,
  pp. 1--8.
\newblock \href {http://dx.doi.org/10.1109/NRSC.2008.4542383}
  {\path{doi:10.1109/NRSC.2008.4542383}}.

\bibitem{bibli17}
P.~H. Abreu, D.~Hugo~Amaro, Castro~Silva, M.~H.~A. Penousal~Machado, N.~Afonso,
  A.~Dourado, Overall survival prediction for women breast cancer using
  ensemble methods and incomplete clinical data, 2014.
\newblock \href {http://dx.doi.org/10.1007/978-3-319-00846-2_338}
  {\path{doi:10.1007/978-3-319-00846-2_338}}.

\bibitem{bibli18}
S.~G. Jacob, R.~G. Ramani, Efficient classifier for classification of
  prognostic breast cancer data through data mining techniques, in: Proceedings
  of the World Congress on Engineering and Computer Science, Vol. (1), 2012,
  pp. 24--26.

\bibitem{bibli6}
B.~S. Jensen~PB, Jensen~LJ, Mining electronic health records: towards better
  research applications and clinical care, Nature Reviews Genetics (13).

\bibitem{bibli7}
R.~D. Kush, E.~Helton, F.~W. Rockhold, C.~D. Hardison, Electronic health
  records, medical research, and the tower of babel, New England Journal of
  Medicine (358)~(16) (2008) 1738--1740.
\newblock \href {http://dx.doi.org/10.1056/NEJMsb0800209}
  {\path{doi:10.1056/NEJMsb0800209}}.

\bibitem{bibli8}
P.~Taylor, Personal genomes: When consent gets in the way, Nature (456).

\bibitem{bibli15}
W.~F.~S. Jionglin~Wu, Jason~Roy, Prediction modeling using ehr data:
  Challenges, strategies, and a comparison of machine learning approaches,
  Medical Care (48)~(6) (2010) S106--S113.

\bibitem{bibli19}
R.~G. Ramani, S.~G. Jacob, Improved classification of lung cancer tumors based
  on structural and physicochemical properties of proteins using data mining
  models, PloS one (8)~(3) (2013) e58772.

\bibitem{bibli20}
K.~I. Penny, G.~D. Smith, The use of data-mining to identify indicators of
  health related quality of life in patients with irritable bowel syndrome, in:
  Information Technology Interfaces, 2009. ITI '09. Proceedings of the ITI 2009
  31st International Conference on, 2009, pp. 87--92.
\newblock \href {http://dx.doi.org/10.1109/ITI.2009.5196059}
  {\path{doi:10.1109/ITI.2009.5196059}}.

\bibitem{bibli21}
B.~Leke-Betechuoh, T.~Marwala, T.~Tim, M.~Lagazio, Prediction of hiv status
  from demographic data using neural networks, in: 2006 IEEE International
  Conference on Systems, Man and Cybernetics, Vol. (3), 2006, pp. 2339--2344.
\newblock \href {http://dx.doi.org/10.1109/ICSMC.2006.385212}
  {\path{doi:10.1109/ICSMC.2006.385212}}.

\bibitem{bibli22}
Z.~A. Altikardes, H.~Erdal, A.~F. Baba, H.~Tezcan, A.~S. Fak, H.~Korkmaz, A
  study to classify non-dipper/dipper blood pressure pattern of type 2 diabetes
  mellitus patients without holter device, in: Computer Applications and
  Information Systems (WCCAIS), 2014 World Congress on, 2014, pp. 1--5.
\newblock \href {http://dx.doi.org/10.1109/WCCAIS.2014.6916555}
  {\path{doi:10.1109/WCCAIS.2014.6916555}}.

\bibitem{bibli23}
D.~Raju, X.~Su, P.~A. Patrician, L.~A. Loan, M.~S. McCarthy, Exploring factors
  associated with pressure ulcers: A data mining approach, International
  Journal of Nursing Studies (52) (2015) 102--111.
\newblock \href {http://dx.doi.org/10.1016/j.ijnurstu.2014.08.002}
  {\path{doi:10.1016/j.ijnurstu.2014.08.002}}.

\bibitem{bibli30}
W.~Dai, T.~S. Brisimi, W.~G. Adams, T.~Mela, V.~Saligrama, I.~C. Paschalidis,
  Prediction of hospitalization due to heart diseases by supervised learning
  methods, International Journal of Medical Informatics (84)~(3) (2015)
  189--197.
\newblock \href {http://dx.doi.org/dx.doi.org/10.1016/j.ijmedinf.2014.10.002}
  {\path{doi:dx.doi.org/10.1016/j.ijmedinf.2014.10.002}}.

\bibitem{bibli31}
T.~Fawcett, An introduction to roc analysis, Pattern Recognition Letter
  (27)~(8) (2006) 861--874.
\newblock \href {http://dx.doi.org/10.1016/j.patrec.2005.10.010}
  {\path{doi:10.1016/j.patrec.2005.10.010}}.

\bibitem{bibli32}
J.~A. Swets, Signal Detection Theory and Roc Analysis in Psychology and
  Diagnostics: Collected Papers, MLawrence Erlbaum Associates, 1996.

\bibitem{bibli33}
T.~Hastie, R.~Tibshirani, J.~Friedman, The Elements of Statistical Learning,
  2nd Edition, Springer, 2009.
\newblock \href {http://dx.doi.org/10.1007/978-0-387-84858-7}
  {\path{doi:10.1007/978-0-387-84858-7}}.

\bibitem{zhu2001kernel}
J.~Zhu, T.~Hastie, Kernel logistic regression and the import vector machine,
  in: Advances in neural information processing systems, 2001, pp. 1081--1088.

\bibitem{bibli29}
S.~Le~Cessie, J.~Van~Houwelingen, Ridge estimators in logistic regression,
  Applied Statistics (41) (1992) 191--201.

\bibitem{bibli27}
C.~Cortes, V.~Vapnik, Support-vector networks, Machine Learning (20)~(3) (1995)
  273--297.
\newblock \href {http://dx.doi.org/10.1007/BF00994018}
  {\path{doi:10.1007/BF00994018}}.

\bibitem{bibli28}
B.~Scholkopf, K.-K. Sung, C.~J.~C. Burges, F.~Girosi, P.~Niyogi, T.~Poggio,
  V.~Vapnik, Comparing support vector machines with gaussian kernels to radial
  basis function classifiers, IEEE Transactions on Signal Processing (45)~(11)
  (1997) 2758--2765.
\newblock \href {http://dx.doi.org/10.1109/78.650102}
  {\path{doi:10.1109/78.650102}}.

\bibitem{bibli24}
H.~Zhang, The optimality of naive bayes, in: Proceedings of the Seventeenth
  International Florida Artificial Intelligence Research Society Conference,
  Miami Beach, Florida, USA, 2004.

\bibitem{bibli25}
L.~Breiman, Random forests, Maching Learning (45)~(1) (2001) 5--32.
\newblock \href {http://dx.doi.org/10.1023/A:1010933404324}
  {\path{doi:10.1023/A:1010933404324}}.

\bibitem{bibli26}
G.~Biau, Analysis of a random forests model, Journal of Machine Learning
  Research (13)~(1) (2012) 1063--1095.

\bibitem{weka}
M.~Hall, E.~Frank, G.~Holmes, B.~Pfahringer, P.~Reutemann, I.~H. Witten, The
  weka data mining software: An update, SIGKDD Explorations (11) (2009)
  1871--1874, software available at
  http://www.csie.ntu.edu.tw/~cjlin/liblinear.

\bibitem{liblin}
R.-E. Fan, K.-W. Chang, C.-J. Hsieh, X.-R. Wang, C.-J. Lin, {LIBLINEAR}: A
  library for large linear classification, Journal of Machine Learning Research
  (9) (2008) 1871--1874, software available at
  http://www.csie.ntu.edu.tw/~cjlin/liblinear.

\bibitem{smolabook}
B.~Sch{\"o}lkopf, A.~J. Smola, Learning with kernels: support vector machines,
  regularization, optimization, and beyond, MIT press, 2002.

\bibitem{chawla2002smote}
N.~V. Chawla, K.~W. Bowyer, L.~O. Hall, W.~P. Kegelmeyer, Smote: synthetic
  minority over-sampling technique, Journal of artificial intelligence research
  16 (2002) 321--357.

\bibitem{libsvm}
C.-C. Chang, C.-J. Lin, Libsvm: A library for support vector machines, ACM
  Transactions on Intelligent Systems and Technology 2 (2011) 27:1--27:27.

\bibitem{bibli34}
Medifee, \href{http://www.medifee.com/}{Medical price search}, [Online;
  accessed 19-Mar-2015].
\newline\urlprefix\url{http://www.medifee.com/}

\bibitem{bibli35}
D.~P.~P. Labs, \href{http://www.phadkelabs.com/price_calculator.php}{Price
  calculator}, [Online; accessed 19-Mar-2015].
\newline\urlprefix\url{http://www.phadkelabs.com/price_calculator.php}

\end{thebibliography}

\end{document}